\documentclass[pdflatex,iicol]{sn-jnl}

\usepackage{graphicx}%
\usepackage{multirow}%
\usepackage{amsmath,amssymb,amsfonts}%
\usepackage{amsthm}%
\usepackage{mathrsfs}%
\usepackage[title]{appendix}%
\usepackage{xcolor}%
\usepackage{textcomp}%
\usepackage{manyfoot}%
\usepackage{booktabs}%
\usepackage{algorithm}%
\usepackage{algorithmicx}%
\usepackage{algpseudocode}%
\usepackage{listings}%

\begin{document}

\title[Article Title]{Pediatric Wrist Fracture Detection Using Feature Context Excitation Modules in X-ray Images}

\author[1]{\fnm{Rui-Yang} \sur{Ju}}\email{jryjry1094791442@gmail.com}
\author[2]{\fnm{Chun-Tse} \sur{Chien}}\email{popper0927@hotmail.com}
\author[3]{\fnm{Enkaer} \sur{Xieerke}}\email{enkaer@mail.shiep.edu.cn}
\author*[2]{\fnm{Jen-Shiun} \sur{Chiang}}\email{jsken.chiang@gmail.com}

\affil*[1]{\orgdiv{Graduate Institute of Networking and Multimedia}, \orgname{National Taiwan University}, \orgaddress{\city{Taipei City}, \postcode{106319}, \country{Taiwan}}}

\affil[2]{\orgdiv{Department of Electrical and Computer Engineering}, \orgname{Tamkang University}, \orgaddress{\city{New Taipei City}, \postcode{251301}, \country{Taiwan}}}

\affil[3]{\orgdiv{College of Energy and Mechanical Engineering}, \orgname{Shanghai University of Electric Power}, \orgaddress{\city{Shanghai}, \postcode{201306}, \country{China}}}

\abstract{Children often suffer wrist trauma in daily life, while they usually need radiologists to analyze and interpret X-ray images before surgical treatment by surgeons.
The development of deep learning has enabled neural networks to serve as computer-assisted diagnosis (CAD) tools to help doctors and experts in medical image diagnostics.
Since YOLOv8 model has obtained the satisfactory success in object detection tasks, it has been applied to various fracture detection.
This work introduces four variants of Feature Contexts Excitation-YOLOv8 (FCE-YOLOv8) model, each incorporating a different FCE module (i.e., modules of Squeeze-and-Excitation (SE), Global Context (GC), Gather-Excite (GE), and Gaussian Context Transformer (GCT)) to enhance the model performance.
Experimental results on GRAZPEDWRI-DX dataset demonstrate that our proposed YOLOv8+GC-M3 model improves the mAP@50 value from 65.78\% to 66.32\%, outperforming the state-of-the-art (SOTA) model while reducing inference time.
Furthermore, our proposed YOLOv8+SE-M3 model achieves the highest mAP@50 value of 67.07\%, exceeding the SOTA performance.
The implementation of this work is available at \url{https://github.com/RuiyangJu/FCE-YOLOv8}.}




\maketitle

\section{Intorduction}
Wrist trauma is common among children~\cite{randsborg2013fractures}.
If they are not treated promptly and effectively, such trauma will result in wrist joint deformities, limited joint motion, and chronic pain~\cite{bamford2010qualitative}.
In severe cases, an incorrect diagnosis would lead to lifelong complications and inconvenience~\cite{mounts2011most}.

With the development of deep learning, neural networks are increasingly used as computer-assisted diagnosis (CAD) tools to aid doctors and experts in analyzing medical images.
Object detection networks can accurately predict fractures and reduce the probability of misdiagnosis.
Although two-stage object detection networks obtain the excellent performance, one-stage object detection networks offer faster inference time.
With the continued release of You Only Look Once (YOLO) series models, YOLOv8~\cite{jocher2023yolo} and YOLOv9~\cite{wang2024yolov9} models can perform object detection efficiently on low-computing platforms.

Since the release of the GRAZPEDWRI-DX~\cite{nagy2022pediatric} dataset by the Medical University of Graz, Ju \emph{et al.}~\cite{ju2023fracture} have been the first to deploy YOLOv8 models to predict pediatric wrist fractures.
This deployment helped surgeons to interpret fractures in X-ray images, reducing misdiagnosis and providing a better information base for the surgery.
Chien \emph{et al.}~\cite{chien2024yolov8} developed YOLOv8-AM models by incorporating different attention modules into YOLOv8~\cite{jocher2023yolo}, which improved the model performance.
In addition, Chien \emph{et al.}~\cite{chien2024yolov9} applied YOLOv9 to this dataset, achieving the state-of-the-art (SOTA) model performance.
However, there remains a great possibility for improving the performance of current SOTA models.

\begin{figure}[t]
\centering
\includegraphics[width=\columnwidth]{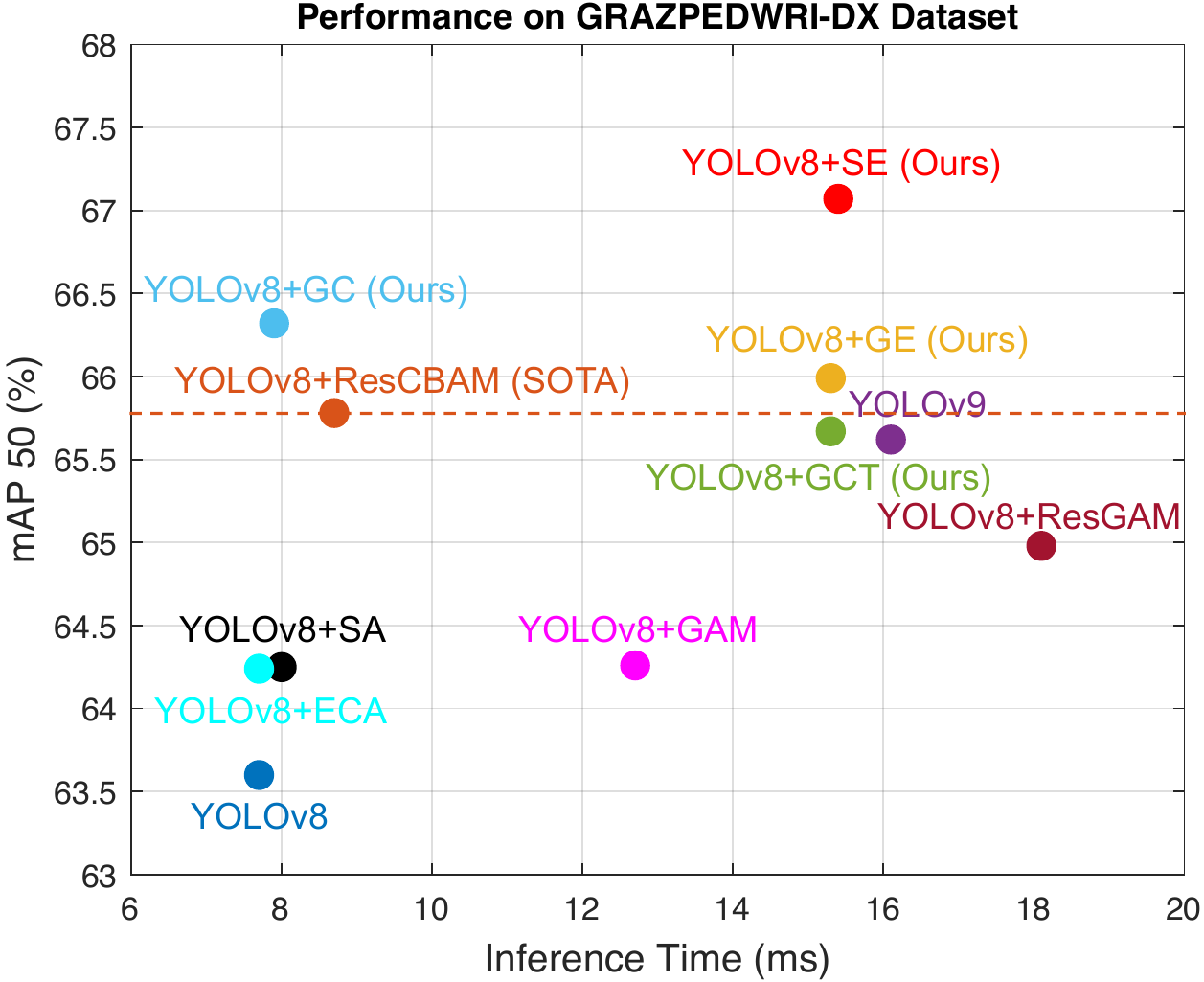}
\setlength{\tabcolsep}{0.8em}{
\begin{tabular}{lcc}
\toprule
\textbf{Module} & \textbf{mAP$_{50}\rm ^{val}$}& \textbf{Inference} \\ \midrule
ResCBAM (SOTA) & 65.78\% & 8.7ms \\ 
GC-M3 (Ours) & 66.32\% & 7.9ms \\ 
GCT-M2 (Ours) & 65.67\% & 15.3ms \\ 
SE-M3 (Ours) & 67.07\% & 15.3ms \\
GE-M2 (Ours) & 65.99\% & 15.3ms \\ \bottomrule
\end{tabular}}
\caption{\textbf{GRAZPEDWRI-DX Dataset mAP@50 vs. Inference Time.} The input image size is 1024; the sizes of YOLOv8 and its improved models are all large (L), and the size of YOLOv9 model is extended (E).}
\label{figure_intro}
\end{figure}

This work proposes FCE-YOLOv8 models, incorporating SE~\cite{hu2018squeeze}, GC~\cite{cao2020global}, GE~\cite{hu2018gather}, and GCT~\cite{ruan2021gaussian} modules into YOLOv8, respectively, to enhance the model performance, and the results are shown in Fig.~\ref{figure_intro}.
This paper makes the following contributions:
\textbf{1) Introduces four different FCE modules to YOLOv8 model architecture, demonstrating significant improvements in model performance for pediatric wrist fracture detection.}
\textbf{2) Evaluates the most efficient model architecture method for each variant of the FCE-YOLOv8 models.}
\textbf{3) Our proposed YOLO+GC model achieves the second-best performance on mAP@50 value on the GRAZPEDWRI-DX public dataset with only a slight increase in inference time.}

The rest of this paper is organized as follows: 
Section~\ref{related} reviews previous research on pediatric wrist fracture detection based on YOLO models.
Section~\ref{method} introduces FCE-YOLOv8 model architectures on three different improved methods, and FCE modules used in this work.
Section~\ref{experiments} presents the experiment conducted to evaluate the most efficient model architecture method for FCE-YOLOv8 models, and compares our proposed models with other SOTA models.
Section~\ref{discussion} discusses the limitations of this work, particularly the class imbalance in the used dataset.
Finally, Section~\ref{conclusion} summaries the outcomes of this paper and suggests directions for future work.

\begin{figure*}[ht]
\centering
\includegraphics[width=\linewidth]{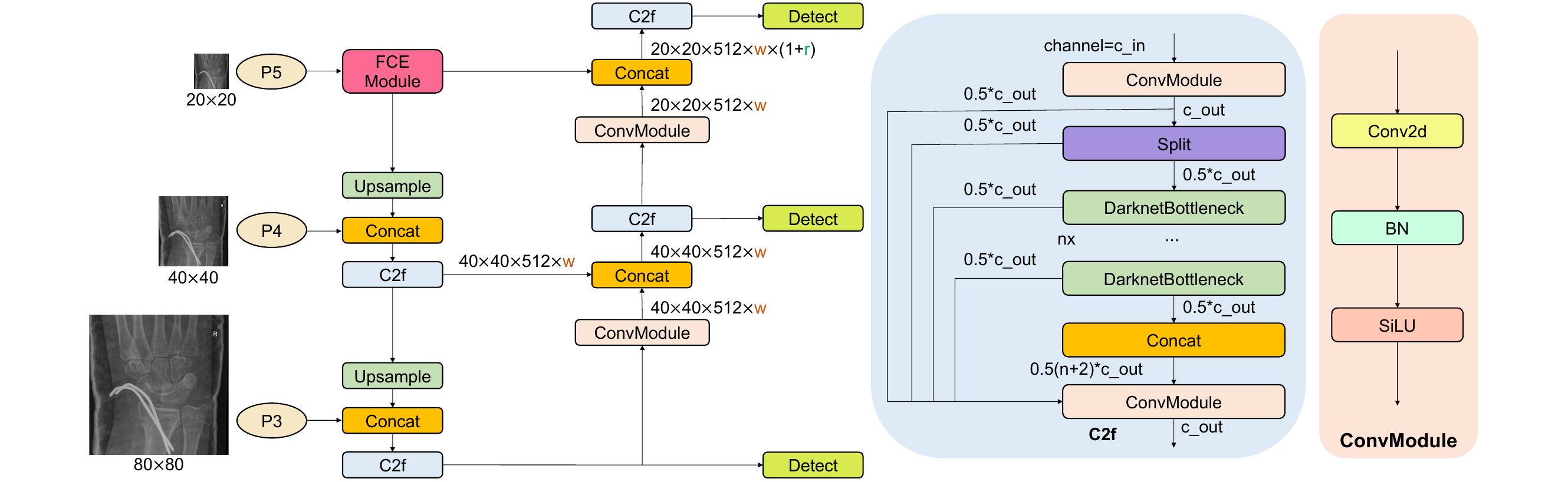}
\caption{Network architecture of improved method-1 (M1): adding one FCE module to the Backbone component of YOLOv8.}
\label{figure_arch_1}
\end{figure*}

\begin{figure*}[ht]
\centering
\includegraphics[width=\linewidth]{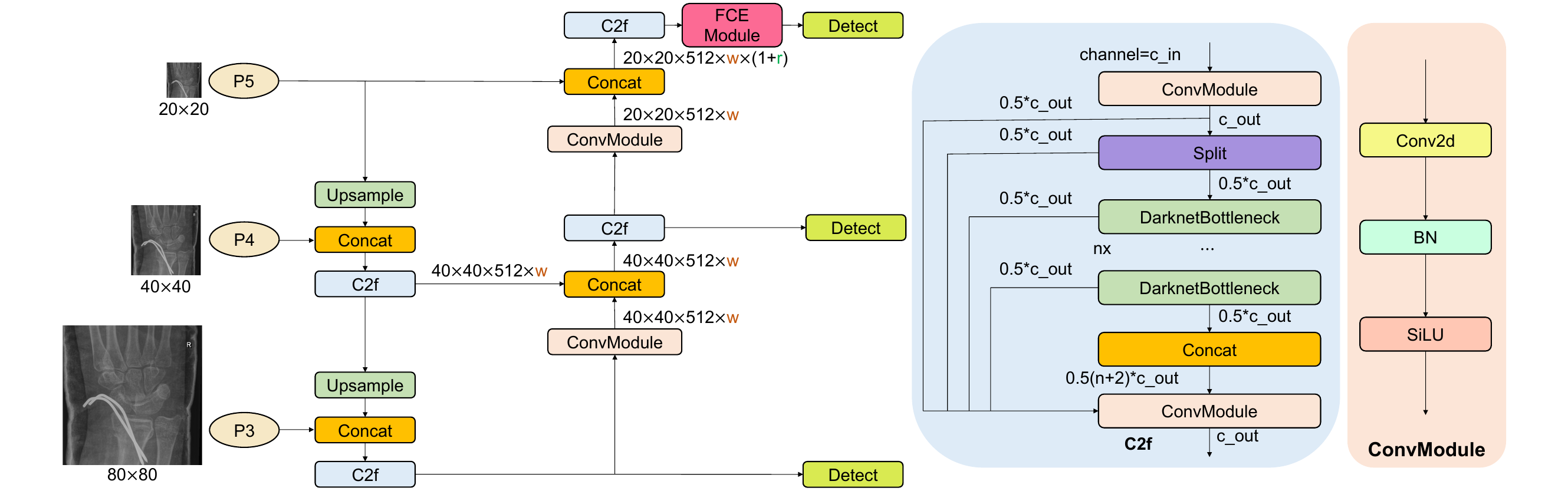}
\caption{Network architecture of improved method-2 (M2): adding one FCE module to the Head component of YOLOv8.}
\label{figure_arch_2}
\end{figure*}

\section{Related Work}
\label{related}
Pediatric wrist fracture detection is an important topic in medical image diagnostic tasks.
Publicly available datasets~\cite{rajpurkar2017mura} for wrist fracture detection in adults are limited, and even fewer datasets~\cite{halabi2019rsna} for pediatrics.
Nagy \emph{et al}~\cite{nagy2022pediatric} published a pediatric related dataset named the GRAZPEDWRI-DX dataset.
Hržić \emph{et al.}~\cite{hrvzic2022fracture} demonstrated that YOLOv4~\cite{bochkovskiy2020yolov4} can enhance the accuracy of pediatric wrist trauma diagnoses in X-ray images.
Ju \emph{et al.}~\cite{ju2023fracture} developed an application employing YOLOv8 to assist doctors in interpreting pediatric wrist tramua X-ray images.
Chien \emph{et al.}~\cite{chien2024yolov8} introduced four attention modules, including Convolutional Block Attention Module (CBAM)~\cite{woo2018cbam}, Efficient Channel Attention (ECA)~\cite{wang2020eca}, Global Attention Mechanism (GAM)~\cite{liu2021global}, and Shuffle Attention (SA)~\cite{zhang2021sa} to improve the model performance, where YOLOv8+ResCBAM achieving the SOTA performance on the mAP@50 value.
In addition, Chien \emph{et al.}~\cite{chien2024yolov9} applied YOLOv9~\cite{wang2024yolov9} to this dataset, and achieved excellent performance.


\begin{figure*}[ht]
\centering
\includegraphics[width=\linewidth]{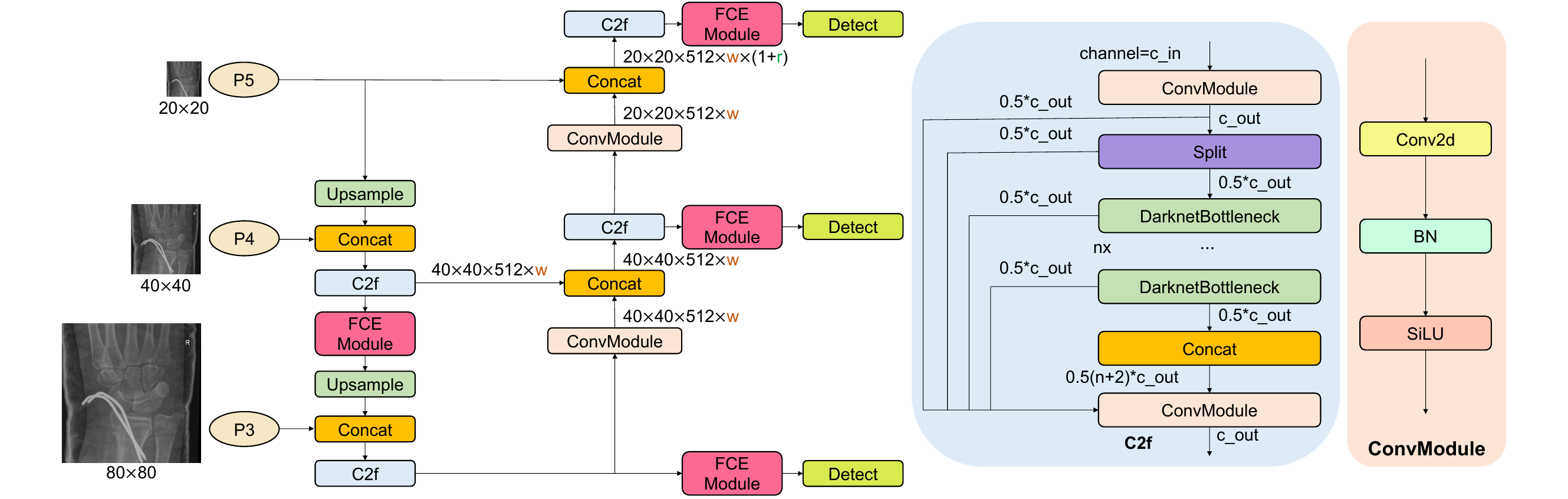}
\caption{Network architecture of improved method-3 (M3): adding four FCE modules to the Head component of YOLOv8.}
\label{figure_arch_3}
\vspace{-0.5em}
\end{figure*}

\begin{figure*}[ht]
\centering
\includegraphics[width=\linewidth]{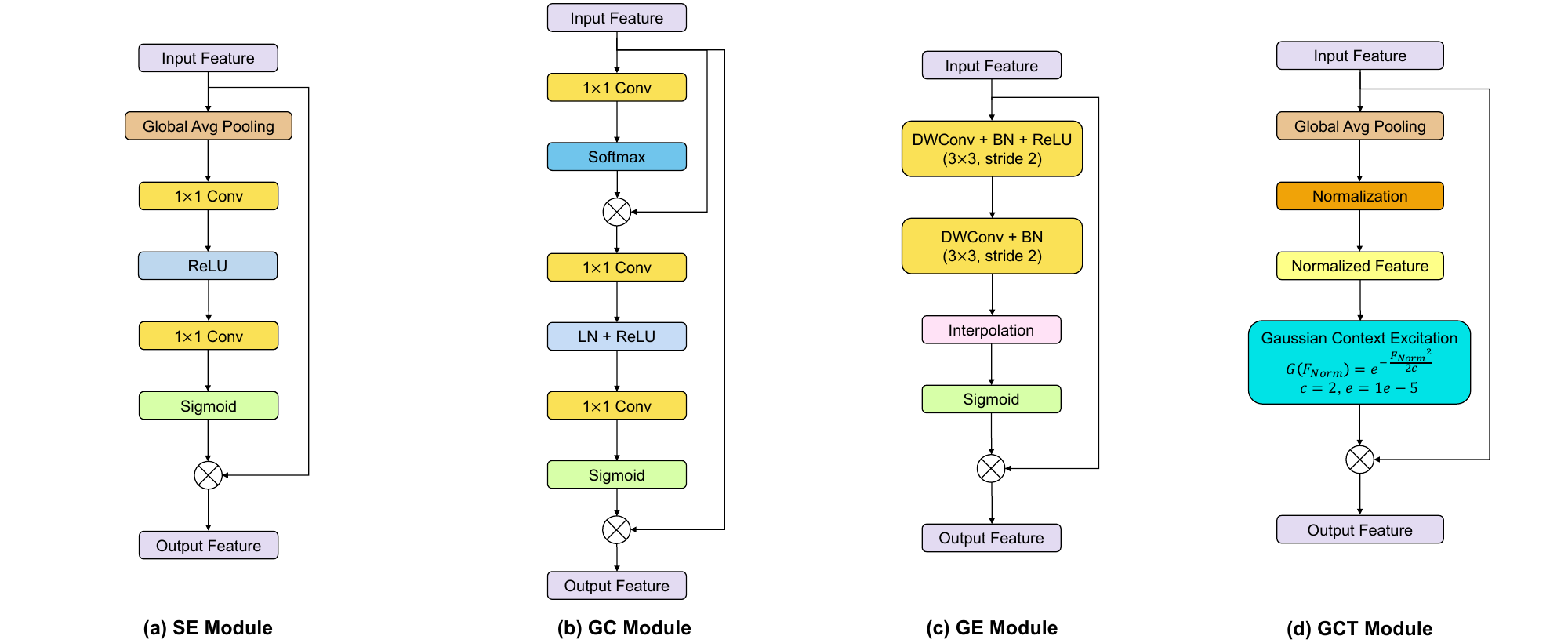}
\caption{Diagram of proposed FCE modules: (a) SE module, (b) GC module, (c) GE module, and (d) GCT module.}
\label{figure_module}
\end{figure*}

\section{Proposed Method}
\label{method}
\subsection{Baseline Model}
\label{architecture}
The network architecture of YOLOv8~\cite{jocher2023yolo} comprises three components: Backbone, Neck, and Head.
The Backbone component employs the Cross Stage Partial (CSP)~\cite{wang2020cspnet} strategy, which divides the feature map into two parts.
One part uses convolution operations, while the other is concatenated with the convolutional output of the first part.
In the Neck component, YOLOv8 applies a combination of Feature Pyramid Network (FPN) and Path Aggregation Network (PAN) for multi-scale feature fusion.
This component deletes convolution operations during the upsampling stage, allowing the top feature maps to retain more information by utilizing more layers, while minimizing the loss of location information in the bottom feature maps due to fewer convolutional layers.

To enhance global feature extraction and capture comprehensive information from medical images, we have designed three improved methods by incorporating different FCE modules into different components of YOLOv8 network architecture, as illustrated in Figs.~\ref{figure_arch_1}, \ref{figure_arch_2}, and \ref{figure_arch_3}, respectively.
Specifically, in the improved method 1 (M1), FCE modules (i.e., SE, GC, GE, and GCT) are added after the Spatial Pyramid Pooling - Fast (SPPF)~\cite{jocher2023yolo} layer in the Backbone component.
In the improved method 2 (M2), different FCE modules are integrated into the final C2f~\cite{jocher2023yolo} module of the Head component.
While in the improved method 3 (M3), FCE modules are added after each of the four C2f modules in the Head component.

\subsection{FCE Modules}
\textbf{SE} enhances the representational capacity of neural networks, including YOLOv8, by dynamically adjusting the weights of feature channels based on their importance. 
This enables YOLOv8 to focus more on relevant channel features while disregarding less significant ones.
Specifically, SE employs the ``Squeeze'' and "Excitation" operations to enhance the representation of meaningful features, improving both classification and detection performance. 
In object detection tasks (e.g., fracture detection), integrating SE into YOLOv8 can significantly improve the model performance

\textbf{GC} is highly suitable for fracture detection tasks due to its flexible properties and ease of integration into different neural network architectures, including the YOLOv8 model.
This module combines the advantages of the Simplified Non-Local (SNL)~\cite{cao2020global} block, which effectively models long-range dependencies, and the SE module, known for its computational efficiency.
In addition, GC follows the NL~\cite{wang2018non} block and aggregates the global context across all locations, enabling the capture of long-range dependencies.
Furthermore, GC is lightweight and can be applied to multiple layers, enhancing the model ability to capture long-range dependencies with only a slight increase in computational cost.

\textbf{GE} can efficiently aggregate the global context information while maintaining spatial resolution, without significantly increasing the model computational demands. 
This makes it possible for use in CAD tools on the low-computing-power platform for fracture prediction. 
By gathering the global context information from the entire feature map, GE captures broader context effectively. 
In addition, different from other methods~\cite{huang2019ccnet} that aggregate global context information, GE preserves spatial resolution, which is useful for tasks requiring precise localization, such as fracture detection. 
Overall, GE can enhance the feature representation capabilities of the network, improving YOLOv8 model performance.

\textbf{GCT} can enhance the ability of YOLOv8 to capture context information by incorporating the Gaussian filter. 
This integration improves the capacity of the model for data modeling and allows for more effective utilization of training data by weighting both local and global context information. 
In addition, this module increases the efficiency of data utilization, reducing the demands for extensive datasets. 
Furthermore, by replacing the traditional self-attention mechanism with the Gaussian filter, GCT decreases computational complexity and enhances efficiency, particularly with large-scale data. 
In general, the Gaussian filter contributes to the model stability during training, mitigating issues such as gradient explosion or vanishing gradients, which results in improved convergence model speed and performance.

\section{Experiments}
\label{experiments}
\subsection{Datasets}
GRAZPEDWRI-DX dataset, introduced by Nagy \emph{et al.}~\cite{nagy2022pediatric}, is a publicly available collection of pediatric wrist trauma X-ray images provided by the Medical University of Graz. 
This dataset was collected by radiologists at the University Hospital Graz between 2008 and 2018, and includes 20,327 X-ray images.
In addition, it covers 6,091 pediatric patients and contains 10,643 studies, with a total of 74,459 image labels and 67,771 annotated objects.
Nagy \emph{et al.}~\cite{nagy2022pediatric} emphasize that prior to the release of their dataset, there are few publicly available relevant pediatric datasets.
Therefore, we only used this dataset for our experiments.

\subsection{Evaluation Metrics}
\label{metric}
Fracture detection on pediatric wrist X-ray images belongs to object detection tasks, and six evaluation metrics in object detection tasks are applicable, including the model parameters (Params), floating-point operations (FLOPs), F1-Score, mean average precision at 50\% intersection over union (mAP@50), mean average precision from 50\% to 95\% intersection over union (mAP@50-95), and the inference time per image.

\textbf{Params} in the model are determined by the architecture layer count, the number of neurons per layer, and the overall complexity of the model.
In general, an increase in Params corresponds to a larger model size.
While larger models generally offer better model performance, they also require more computational resources.
Therefore, it is important to balance the model size with computational cost in practical applications.

\textbf{FLOPs} representing floating-point operations, are commonly used to evaluate the computational complexity of the model.
This metric serves as an indicator of the computational performance and speed of the models.
In general, the models with lower FLOPs are more suitable for resource-limited environments, while the models with higher FLOPs require more powerful hardware support.

\textbf{F1-Score} is the harmonic mean of the Precision and Recall of the model, ranging from 0 to 1.
The value closer to 1 indicates that the model has achieved a better balance between Precision and Recall.
Conversely, if either Precision or Recall is close to 0, F1-Score will also approach 0, signaling poor model performance.

\textbf{mAP} is a widely used metric for evaluating the performance of object detection models.
In object detection tasks, the model aims to recognize objects within an image and determines their locations.
For each class, the model calculates the area under the Precision-Recall curve, known as average precision (AP), which measures the model detection performance for that specific class.
The mAP score is calculated by averaging the AP values across all classes, providing an overall measure of the model detection accuracy.

\textbf{Inference time} denotes the time taken by the network model from the input of an X-ray image to the generation of the prediction output, encompassing preprocessing, inference, and post-processing steps. In this work, inference time per image was measured using a single NVIDIA GeForce RTX 3090 GPU.

\begin{table}[t]
\caption{Experimental results of the \textbf{SE module} based on YOLOv8 on the GRAZPEDWRI-DX dataset. $^1$The input image size is 640. $^2$The input image size is 1024.}
\setlength{\tabcolsep}{0.5em}{
\begin{tabular}{lcccc}
\toprule
\textbf{Module} & \textbf{Params} & \textbf{FLOPs} & \textbf{mAP$_{50}\rm ^{val}$} & \textbf{Inference} \\ \midrule
SE-M1-S & 11.30M & 29.1G & 62.18\% & 2.6ms \\
SE-M1-M & 26.12M & 79.8G & 62.35\% & 5.1ms \\
SE-M1-L & 43.93M & 166.3G & 62.93\% & 7.6ms \\ \midrule
SE-M2-S & 11.17M & 28.7G & 61.45\% & 2.5ms \\
SE-M2-M & 25.90M & 79.1G & 61.75\% & 5.0ms \\
SE-M2-L & 43.67M & 165.5G & 64.08\% & 8.0ms \\ \midrule
SE-M3-S & 11.19M & 28.7G & 61.72\% & 2.6ms \\
SE-M3-M & 25.95M & 79.2G & 62.63\% & 5.0ms \\
SE-M3-L & 43.85M & 165.6G & 62.79\% & 7.6ms \\ \midrule \midrule
SE-M1-S & 11.30M & 29.1G & 63.24\% & 4.9ms \\
SE-M1-M & 26.12M & 79.8G & 63.70\% & 10.1ms \\
SE-M1-L & 43.93M & 166.3G & 64.55\% & 15.4ms \\ \midrule
SE-M2-S & 11.17M & 28.7G & 62.90\% & 5.0ms \\
SE-M2-M & 25.90M & 79.1G & 63.68\% & 10.0ms \\
SE-M2-L & 43.67M & 165.5G & 64.40\% & 14.7ms \\ \midrule
SE-M3-S & 11.19M & 28.7G & 62.77\% & 5.1ms \\
SE-M3-M & 25.95M & 79.2G & 64.10\% & 10.0ms \\
SE-M3-L & 43.85M & 165.6G & 67.07\% & 15.3ms \\ \bottomrule
\end{tabular}}
\label{table_se}
\end{table}

\subsection{Data Preparation}
To ensure fairness in model performance comparison, the same data is used for all model training.
We randomly divide the GRAZPEDWRI-DX~\cite{nagy2022pediatric} dataset into training, validation, and test sets in the ratio of 70\%, 20\%, and 10\%, respectively.
Specifically, the training set contains 14,204 images (69.88\%); the validation set contains 4,094 images (20.14\%), and the test set contains 2,029 images (9.98\%).
Furthermore, data augmentation is performed on the training set before all models training.
Specifically, this work adjusts the contrast and brightness of the X-ray images using the addWeighted function in the Open Source Computer Vision Library (OpenCV), and expanding the original training dataset from 14,204 images to 28,408 images.

\begin{table}[t]
\caption{Experimental results of the \textbf{GC module} based on YOLOv8 on the GRAZPEDWRI-DX dataset. $^1$The input image size is 640. $^2$The input image size is 1024.}
\setlength{\tabcolsep}{0.5em}{
\begin{tabular}{lcccc}
\toprule
\textbf{Module} & \textbf{Params} & \textbf{FLOPs} & \textbf{mAP$_{50}\rm ^{val}$} & \textbf{Inference} \\ \midrule
GC-M1-S & 11.34M & 29.1G & 60.68\% & 1.7ms \\
GC-M1-M & 26.17M & 79.9G & 63.07\% & 2.5ms \\
GC-M1-L & 43.97M & 166.3G & 64.55\% & 3.5ms \\ \midrule
GC-M2-S & 11.21M & 28.7G & 61.30\% & 1.9ms \\
GC-M2-M & 25.95M & 79.2G & 62.71\% & 2.7ms \\
GC-M2-L & 43.70M & 165.5G & 64.54\% & 3.5ms \\ \midrule
GC-M3-S & 11.24M & 28.7G & 62.19\% & 1.9ms \\
GC-M3-M & 26.03M & 79.2G & 62.77\% & 2.7ms \\
GC-M3-L & 43.85M & 165.6G & 63.97\% & 3.6ms \\ \midrule \midrule
GC-M1-S & 11.34M & 29.1G & 63.24\% & 2.9ms \\
GC-M1-M & 26.17M & 79.9G & 63.87\% & 5.1ms \\
GC-M1-L & 43.97M & 166.3G & 64.85\% & 7.8ms \\ \midrule
GC-M2-S & 11.21M & 28.7G & 63.51\% & 2.8ms \\
GC-M2-M & 25.95M & 79.2G & 65.25\% & 5.1ms \\
GC-M2-L & 43.70M & 165.5G & 65.70\% & 7.8ms \\ \midrule
GC-M3-S & 11.24M & 28.7G & 64.16\% & 2.8ms \\ 
GC-M3-M & 26.03M & 79.2G & 64.33\% & 5.2ms \\
GC-M3-L & 43.85M & 165.6G & 66.32\% & 7.9ms \\ \bottomrule
\end{tabular}}
\label{table_gc}
\end{table}

\subsection{Training}
To ensure a fair comparison of the performance of different models on the GRAZPEDWRI-DX~\cite{nagy2022pediatric} dataset for pediatric wrist fracture detection, the same training environment and hardware (NVIDIA GeForce RTX 3090 GPUs) are used for all models training and evaluation.
This work employs Python 3.9 and the PyTorch framework for training all of the models.
For the hyperparameters of the model training, we set the batch size to 16 and the number of epochs of 100.
In addition, all models are trained using the SGD optimizer, with the weight decay of 0.0005, momentum of 0.937, and the initial learning rate of 0.01.

\begin{table}[t]
\caption{Experimental results of the \textbf{GE module} based on YOLOv8 on the GRAZPEDWRI-DX dataset. $^1$The input image size is 640. $^2$The input image size is 1024.}
\setlength{\tabcolsep}{0.5em}{
\begin{tabular}{lcccc}
\toprule
\textbf{Module} & \textbf{Params} & \textbf{FLOPs} & \textbf{mAP$_{50}\rm ^{val}$} & \textbf{Inference} \\ \midrule
GE-M1-S & 11.28M & 29.1G & 62.44\% & 2.6ms \\
GE-M1-M & 26.10M & 79.8G & 63.00\% & 5.1ms \\
GE-M1-L & 43.91M & 166.3G & 64.15\% & 7.6ms \\ \midrule
GE-M2-S & 11.15M & 28.7G & 62.07\% & 2.6ms \\
GE-M2-M & 25.87M & 79.1G & 63.20\% & 5.1ms \\
GE-M2-L & 43.65M & 165.4G & 64.02\% & 7.6ms \\ \midrule
GE-M3-S & 11.16M & 28.7G & 61.89\% & 2.7ms \\
GE-M3-M & 25.90M & 79.1G & 62.22\% & 5.2ms \\
GE-M3-L & 43.68M & 165.5G & 62.42\% & 7.7ms \\ \midrule \midrule
GE-M1-S & 11.28M & 29.1G & 63.72\% & 5.2ms \\
GE-M1-M & 26.10M & 79.8G & 63.81\% & 10.0ms \\
GE-M1-L & 43.91M & 166.3G & 64.34\% & 15.4ms \\ \midrule
GE-M2-S & 11.15M & 28.7G & 62.51\% & 5.1ms \\
GE-M2-M & 25.87M & 79.1G & 64.28\% & 10.0ms \\
GE-M2-L & 43.65M & 165.4G & 65.99\% & 15.3ms \\ \midrule
GE-M3-S & 11.16M & 28.7G & 62.42\% & 4.8ms \\ 
GE-M3-M & 25.90M & 79.1G & 62.63\% & 10.3ms \\
GE-M3-L & 43.68M & 165.5G & 64.35\% & 15.7ms \\ \bottomrule
\end{tabular}}
\label{table_ge}
\end{table}

\subsection{Selection of Improved Methods}
\label{selection}
This work presents three different methods for improving model architecture, resulting in FCE-YOLOv8 models. 
Given four proposed FCE modules (i.e., SE, GC, GE, and GCT), we conduct the experiment to determine the most suitable improved method for each module.
The experimental results are shown in Tables~\ref{table_se},~\ref{table_gc},~\ref{table_ge}, and~\ref{table_gct}. 
As described in Section~\ref{architecture}, we apply three methods (i.e., M1, M2, and M3) to train FCE-YOLOv8 with SE, GC, GE, and GCT modules, respectively. 
Apart from the parameters of module, model size, and input image size, all other settings remain the same for all models.

Table~\ref{table_se} presents the experimental results for YOLOv8+SE. 
It can be seen that for the SE module, when the input image size is 640, choosing M2 to construct the YOLOv8+SE model can get the highest mAP@50 value of 64.08\%; while the input image size is 1024, choosing M3 to construct the YOLOv8+SE model can get the highest mAP@50 value of 67.07\%. 
In addition, with the input size of 640, Tables~\ref{table_gc},~\ref{table_ge}, and~\ref{table_gct} show that the most suitable method for GC, GE, and GCT modules is all M1. 
YOLOv8+GC, YOLOv8+GE, and YOLOv8+GCT with M1 achieve the highest mAP@50 values of 64.55\%, 64.15\%, and 63.70\%, respectively. 
However, when the input size is set to 1024, the most effective method for YOLOv8+GC is M3, resulting in the mAP@50 value of 66.32\%. 
For both YOLOv8+GE and YOLOv8+GCT, M2 is the most effective, obtaining the mAP@50 values of 65.99\% and 65.67\%, respectively.
Since the input image size is set to 1024 for comparisons with the SOTA models, as shown in Section~\ref{experiment}, we select M3, M3, M2, and M2 for SE, GC, GE, and GCT, respectively.

\begin{table}[t]
\caption{Experimental results of \textbf{GCT module} based on YOLOv8 on the GRAZPEDWRI-DX dataset. $^1$The input image size is 640. $^2$The input image size is 1024.}
\setlength{\tabcolsep}{0.40em}{
\begin{tabular}{lcccc}
\toprule
\textbf{Module} & \textbf{Params} & \textbf{FLOPs} & \textbf{mAP$_{50}\rm ^{val}$} & \textbf{Inference} \\ \midrule
GCT-M1-S$^1$ & 11.27M & 29.1G & 61.91\% & 2.7ms \\
GCT-M1-M$^1$ & 26.08M & 79.8G & 62.64\% & 5.1ms \\
GCT-M1-L$^1$ & 43.90M & 166.3G & 63.70\% & 7.6ms \\ \midrule
GCT-M2-S$^1$ & 11.14M & 28.7G & 61.36\% & 2.6ms \\
GCT-M2-M$^1$ & 25.86M & 79.1G & 62.19\% & 5.1ms \\
GCT-M2-L$^1$ & 43.64M & 165.4G & 63.42\% & 7.6ms \\ \midrule
GCT-M3-S$^1$ & 11.14M & 28.7G & 52.07\% & 2.7ms \\
GCT-M3-M$^1$ & 25.86M & 79.1G & 53.60\% & 5.2ms \\
GCT-M3-L$^1$ & 43.64M & 165.4G & 57.88\% & 7.6ms \\ \midrule \midrule
GCT-M1-S$^2$ & 11.27M & 29.1G & 62.49\% & 5.1ms \\
GCT-M1-M$^2$ & 26.08M & 79.8G & 63.47\% & 10.1ms \\
GCT-M1-L$^2$ & 43.90M & 166.3G & 63.87\% & 15.3ms \\ \midrule
GCT-M2-S$^2$ & 11.14M & 28.7G & 63.51\% & 5.0ms \\
GCT-M2-M$^2$ & 25.86M & 79.1G & 64.46\% & 10.1ms \\
GCT-M2-L$^2$ & 43.64M & 165.4G & 65.67\% & 15.3ms \\ \midrule
GCT-M3-S$^2$ & 11.14M & 28.7G & 52.53\% & 5.1ms \\ 
GCT-M3-M$^2$ & 25.86M & 79.1G & 53.70\% & 10.1ms \\
GCT-M3-L$^2$ & 43.64M & 165.4G & 58.15\% & 15.7ms \\ \bottomrule
\end{tabular}}
\label{table_gct}
\end{table}

\begin{table*}[t]
\caption{Quantitative comparison with other models for pediatric wrist fracture detection on the GRAZPEDWRI-DX dataset. The sizes of YOLOv8 and its improved models are all large (L), and the size of YOLOv9 model is extended (E). The best performance and the 2nd best performance are in {\color{red}red} and {\color{blue}blue} colors, respectively.}
\setlength{\tabcolsep}{0.8em}{
\begin{tabular}{lcccccc}
\toprule
\textbf{Model} & \textbf{Input Size} & \textbf{Params} & \textbf{FLOPs} & \textbf{F1-Score} & \textbf{mAP$_{50}\rm ^{val}$} & \textbf{Inference} \\ \midrule
YOLOv8 & 1024 & 43.61M & 164.9G & 0.62 & 63.58\% & {\color{red}7.7ms} \\
YOLOv8+SA & 1024 & 43.64M & 165.4G & 0.63 & 64.25\% & 8.0ms \\
YOLOv8+ECA & 1024 & 43.64M & 165.5G & {\color{blue}0.65} & 64.24\% & {\color{red}7.7ms} \\
YOLOv8+GAM & 1024 & 49.29M & 183.5G & {\color{blue}0.65} & 64.26\% & 12.7ms \\
YOLOv8+ResGAM & 1024 & 49.29M & 183.5G & 0.64 & 64.98\% & 18.1ms \\
YOLOv8+ResCBAM & 1024 & 53.87M & 196.2G & 0.64 & 65.78\% & 8.7ms \\
YOLOv9 & 1024 & 69.42M & 244.9G & {\color{red}0.66} & 65.62\% & 16.1ms \\ \midrule
YOLOv8+GC-M3 & 1024 & 43.85M & 165.6G & {\color{red}0.66} & {\color{blue}66.32\%} & {\color{blue}7.9ms} \\
YOLOv8+GCT-M2 & 1024 & 43.64M & 165.4G & 0.64 & 65.67\% & 15.3ms \\
YOLOv8+SE-M3 & 1024 & 43.85M & 165.6G & {\color{red}0.66} & {\color{red}67.07\%} & 15.3ms \\
YOLOv8+GE-M2 & 1024 & 43.65M & 165.4G & 0.64 & 65.99\% & 15.3ms \\ \bottomrule
\end{tabular}}
\label{table_comparison}
\vspace{-0.5em}
\end{table*}

\begin{figure*}[ht]
\centering
\includegraphics[width=\linewidth]{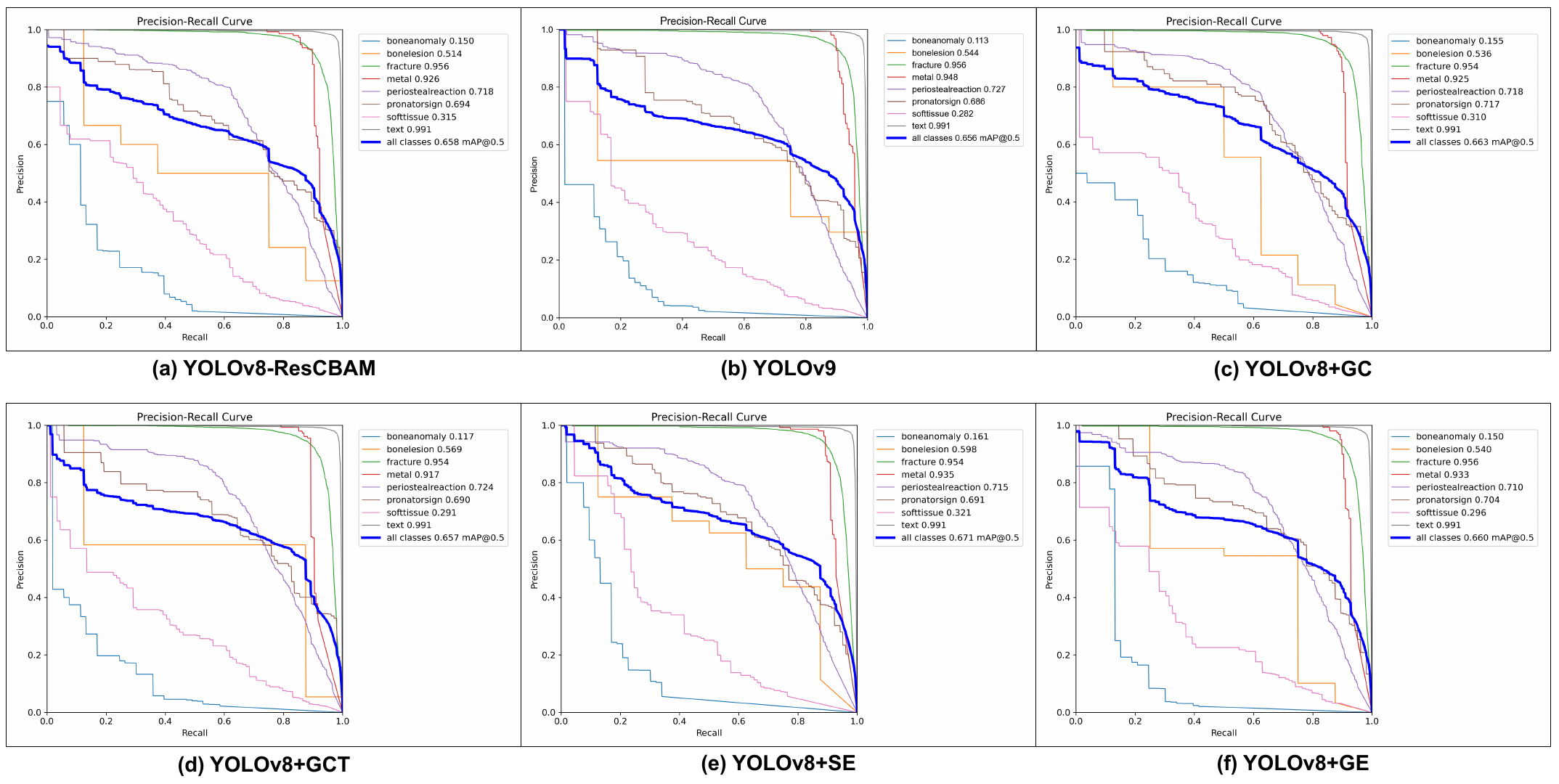}
\vspace{-1em}
\caption{Visualization of the accuracy of predicting each class using our proposed FCE-YOLOv8 models and other SOTA models on the GRAZPEDWRI-DX dataset with the input image size of 1024.
}
\label{figure_result}
\end{figure*}

\begin{figure*}[ht]
\centering
\includegraphics[width=\linewidth]{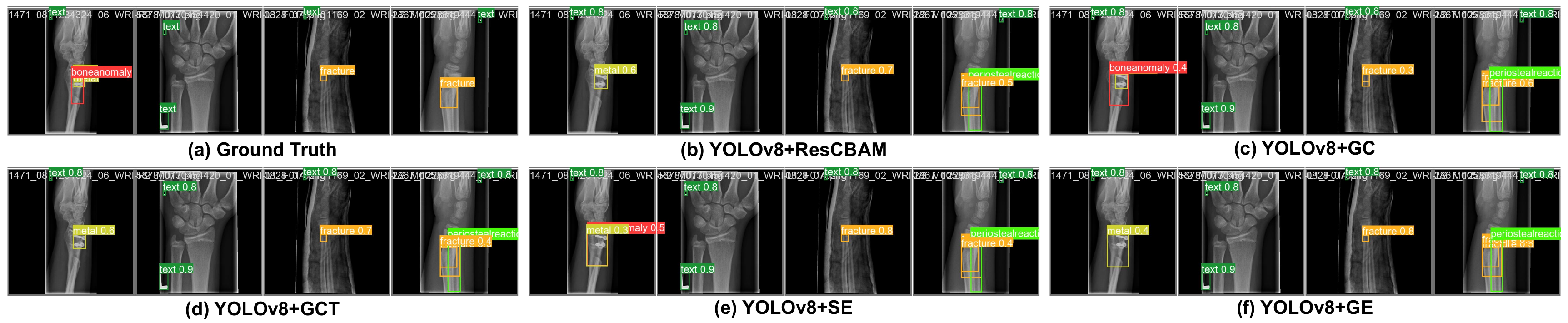}
\vspace{-1em}
\caption{Examples of prediction results by our proposed models and other SOTA models, and Ground-Truth images of pediatric wrist fracture detection on X-ray Images.}
\label{figure_predict}
\end{figure*}

\subsection{Comparison with SOTA Models}
\label{experiment}
For a comparative analysis with other SOTA models, this work sets the input image size to 1024 and configures all YOLOv8 models to large (L) and YOLOv9 to extended (E).
The experimental results in Table~\ref{table_comparison} demonstrate that the proposed FCE-YOLOv8 models achieve satisfactory performance. 
Specifically, our proposed YOLOv8+GC-M3 and YOLOv8+SE-M3 models achieve the same F1-Score as YOLOv9, which is 0.66, reaching the SOTA level.
In addition, the mAP@50 value of our proposed YOLOv8+SE-M3 model obtains the highest value of 67.07\%, surpassing 65.78\% of the previous SOTA model, YOLOv8+ResCBAM.
Furthermore, our YOLOv8+GC-M3 model achieves the second-highest mAP@50 value of 66.32\%.
Based on this, the inference time of our YOLOv8+GC-M3 model is significantly shorter at 7.9 ms, compared to 8.7 ms of YOLOv8+ResCBAM and 16.1 ms of YOLOv9.
For the model parameters and FLOPs, our proposed FCE-YOLOv8 models are also more efficient than the previous SOTA model.
These experimental results demonstrate the positive effect of adding FCE modules to YOLOv8 on the model performance.

Fig.~\ref{figure_result} presents the Precision-Recall Curve of our proposed FCE-YOLOv8 and other SOTA models for each class.
According to Fig.~\ref{figure_result}, all models have good abilities to correctly detect the ``fracture'', ``metal'', and ``text'' classes, with the accuracy all above 90\%.
However, the poor ability of these models to detect the ``bone anomaly'' class seriously affects the mAP@50 values of the models.
The proposed FCE-YOLOv8 models detect the ``bone anomaly'' class with the highest accuracy of 16.1\%, compared to 15.0\% of YOLOv8+ResCBAM, and 11.3\% of YOLOv9.
Details of this issue and its solution will be presented in Section~\ref{discussion}.

In a real-world diagnostic scenario, to evaluate the accuracy improvement of YOLOv8 models enhanced with FCE modules for pediatric wrist fracture detection, four X-ray images are randomly selected from the used dataset.
Fig.~\ref{figure_predict} presents the predictions of our proposed models and YOLOv8+ResCBAM, as well as the Ground-Truth images. 
Compared to the previous SOTA model (i.e., YOLOv8+ResCBAM), our proposed YOLOv8+GC and YOLOv8+SE models demonstrated greater accuracy in detecting ``bone anomalies'', particularly in the first X-ray image on the left side of Fig.~\ref{figure_predict}.
This highlights the suitability of our proposed models as CAD tools to assist doctors and experts.

\section{Discussion}
\label{discussion}
Although the highest mAP@50 value of the proposed FCE-YOLOv8 models achieve 67.07\% (surpassing the SOTA level), it falls short of the 70\%.
This limitation is primarily due to the poor performance of the model in predicting the ``bone anomaly'' and ``soft tissue'' classes.
The GRAZPEDWRI-DX~\cite{nagy2022pediatric} dataset suffers from the class imbalance, for example, compared to 23,722 and 18,090 labels of the ``text'' and ``fracture'' classes, there are only 276 and 464 labels of the ``bone anomaly'' and ``soft tissue'' classes.
The performance of YOLO models mainly depends on the quality and diversity of the training data.
The limited training samples for these underrepresented classes result in the poor prediction ability of the model.
Therefore, to further improve the model performance, obtaining more data for the ``bone anomaly'' and ``soft tissue'' classes is important.
Our research team continues to search for additional data, and given the scarcity of pediatric-related X-ray images, we will immediately apply any newly available datasets to our FCE-YOLOv8 model training.

\section{Conclusion}
\label{conclusion}
Applying YOLOv8~\cite{ju2023fracture}, YOLOv8-AM~\cite{chien2024yolov8,ju2024yolov8}, and YOLOv9~\cite{chien2024yolov9} to pediatric wrist fracture detection has led to significant performance improvements on the publicly available GRAZPEDWRI-DX dataset~\cite{nagy2022pediatric}.
On this dataset, the current SOTA model, YOLOv8+ResCBAM-L, obtains the mAP@50 values of 65.78\%.
However, this model is significantly larger than the original YOLOv8 model in terms of parameters and FLOPs.
For example, the original YOLOv8-L model has only 43.61M parameters, while YOLOv8+ResCBAM-L has 53.87M, which is not satisfactory.
Therefore, this paper proposes FCE-YOLOv8-L to improve the model performance effectively in a lightweight way.
The parameters of all the proposed models are only from 43.64M to 43.85M, while the highest mAP@50 value reaches 67.07\%, which exceeds the SOTA level.

To make this work available as CAD tools for medical image diagnostics, we plan to deploy our proposed FCE-YOLOv8 models as both web application and mobile phone application (e.g., Android and iOS) in the future, ensuring easy accessibility for doctors and experts.

\section{Declarations}
This paper is an expanded paper from International Symposium on Intelligent Signal Processing and Communication Systems (ISPACS) held on December 10-13, 2024 in Kaohsiung, Taiwan.

This work is supported by National Science and Technology Council of Taiwan, under Grant Number: NSTC 112-2221-E-032-037-MY2.

\bibliographystyle{splncs04}
\bibliography{sn-bibliography}
\end{document}